# Data-Driven Gyroscope Calibration

**Zeev Yampolsky**[1] **and Itzik Klein**[1]

[1] The Autonomous Navigation and Sensor Fusion Lab,
The Hatter Department of Marine Technlogies,
University of Haifa,
Haifa, Israel

Abstract

Gyroscopes are inertial sensors that measure the angular velocity of the platforms to which they are attached. To estimate the gyroscope deterministic error terms prior mission start, a calibration procedure is performed. When considering low-cost gyroscopes, the calibration requires a turntable as the gyros are incapable of sensing the Earth turn rate. In this paper, we propose a data-driven framework to estimate the scale factor and bias of a gyroscope. To train and validate our approach, a dataset of 56 minutes was recorded using a turntable. We demonstrated that our proposed approach outperforms the model-based approach, in terms of accuracy and convergence time. Specifically, we improved the scale factor and bias estimation by an average of 72% during six seconds of calibration time, demonstrating an average of 75% calibration time improvement. That is, instead of minutes, our approach requires only several seconds for the calibration.

1. Introduction

An inertial measurement unit (IMU) is comprised of two inertial sensors, an accelerometer which measures the specific force vector, and a gyroscope which measures the angular velocity vector [1]. By integrating both quantities the navigation solution (position velocity, and orientation) can be calculated [2]. Micro-electrical-mechanical-system (MEMS) IMU are low-cost, low-grade, small and compact size IMUs [3]. MEMS IMU are used in a wide range of applications such as quadrotors, human activity recognition, and mobile robots [4], [5], [6].

Although MEMS IMUs wildly applicability, they are considered noisy and inaccurate due to large error terms [7], which in turn results in the navigation solution drift [8]. To mitigate the measurement errors, a stationary calibration procedure is performed to estimate the sensor deterministic error terms [9]. Moreover, when considering low-grade IMUs, the gyroscope errors are more prominent than of the accelerometer [10], [11]. There are two main calibration approaches, a zero-order calibration in which only the biases are estimated, and more comprehensive approaches such as the six positions calibration, in which six error terms are estimated for all three axes. The latter, usually involves rotating the gyroscope on a turntable, at a known angular velocity, to produce a strong single for accurate calibration [7], [9].

Recently, data-driven approaches have been utilized in navigation related domains and specifically in gyroscope related tasks [12], [13], [14]. In [15], machine learning is leveraged to perform gyro-compassing using MEMS-based gyroscopes. In [7], a temporal convolution network is used to estimate the gyroscope output, thus performing model-free calibration. In [11] the authors use a convolution neural network (CNN) to estimate the biases of a single MEMS-based gyroscope, through zero-order calibration. Similarly, in [16], the authors use a CNN to estimate the biases of multiple gyroscopes. Both in [11] and [16] the tradeoff between calibration time and accuracy is extensively discussed, stating that in some application shorter setup time is more important than zero error calibration.

In this research, we aim to further investigate the tradeoff discussed in [11] and [16], and achieve more accurate calibration by estimating the scale factor and the bias simultaneously. To this end, we propose a data-driven approach to estimate both error terms. We employ a 2-dimensional CNN (2DCNN), in an innovative way to estimate both error terms in an end-to-end approach. To evaluate our proposed approach, we recoded a dataset of 56 minutes using gyroscopes mounted on a turntable. We show that our proposed data-driven approach achieves 72% accuracy improvement in estimating the error terms, while shortening the calibration time by up to 75%, compared to a model-based baseline approach.

The rest of this paper is organized as follows, Section 2 describes the gyroscope error models and the model-based calibration approach. Section 3 describes our proposed approach and provides the neural network derivations. Section 4 describes the data collection and shows the results. Finally, Section 5 concludes this research.

## 2. Scientific Background

This section provides the scientific background for the gyroscopes error modelling and model-based gyroscope calibration.

### 2.1. Gyroscope Error Model

In this work, we focus on low-cost, low-performance, gyroscopes. A common gyroscope error model consists of several error terms including scale factor, misalignment, bias, and zero mean white Gaussian noise, modeled by [17], [18]:

$$\widetilde{\boldsymbol{\omega}}^b = \mathbf{M} \cdot \boldsymbol{\omega}^b + \boldsymbol{b}_g + \boldsymbol{w}_g \qquad (1)$$

where $\widetilde{\boldsymbol{\omega}}^b$ is the gyroscope measured angular velocity expressed in the body frame, $\boldsymbol{\omega}^b$ is the true angular velocity signal expressed in the body frame, $\boldsymbol{b}_g$ is an additive error vector referred to as the bias, and $\boldsymbol{w}_g$ is a zero mean white Gaussian noise. The true angular velocity vector, $\boldsymbol{\omega}^b$, is multiplied by the matrix $\mathbf{M}$ which contains the misalignments errors and the scale factor as follows:

$$\mathbf{M} = \mathbf{I}_3 + \mathbf{M}_{SF} + \mathbf{M}_{MA} \tag{2}$$

where $\mathbf{I}_3$ is the identity matrix of size $3 \times 3$, the matrix $\mathbf{M}_{SF}$ is a diagonal matrix of the scale factors applied over each axis of the true angular velocity signal as follows:

$$\mathbf{M}_{SF} = \begin{bmatrix} s_x & 0 & 0 \\ 0 & s_y & 0 \\ 0 & 0 & s_z \end{bmatrix} \tag{3}$$

and $s_x, s_y, s_z$ are multiplicative values referred to as scale factors. The matrix $\mathbf{M}_{MA}$ is the misalignment matrix which reflects the misalignment error between each two axes due to non-orthogonal axes or other installation errors. As such misalignment angles are considered small, in low-cost gyroscopes the bias and scale factor error terms are considered to be more dominant.

Therefore, we neglect the misalignment error terms and thus (1) is reduced to:

$$\tilde{\boldsymbol{\omega}}^b = \begin{bmatrix} 1+s_x & 0 & 0 \\ 0 & 1+s_y & 0 \\ 0 & 0 & 1+s_z \end{bmatrix} \cdot \boldsymbol{\omega}^b + \boldsymbol{b}_g + \boldsymbol{w}_g \tag{4}$$

*2.2.  Model-based Gyroscope Calibration*

One of the most common approaches to perform gyroscope calibration is the six positions calibration approach [19] and its variations, such as the 12 or 24 positions calibration and others [20]. The six position calibration procedure is carried out before the mission starts to estimate the constant error terms of the gyroscope. This approach requires six positions to rotate the gyroscope six times, two times per axis, once clockwise and once counterclockwise. Each rotation lasts for $T$ seconds to later average the measurements and eliminate the sensor noise.

First, to simplify the formulation and derivations of the six positions approach, we rewrite (4) to relate between the true angular velocity and the measured velocity using matrix form as follows:

$$\tilde{\boldsymbol{\omega}}^b = \mathbf{Z} \cdot [\boldsymbol{\omega}^b \quad | \quad 1] + \boldsymbol{w}_g \tag{5}$$

where the term $[\boldsymbol{\omega}^b \mid 1] \in \mathbb{R}^4$ is a vector comprised of the true angular velocity vector stacked with the number one, thus creating a vector with four inputs, the matrix $\mathbf{Z} \in \mathbb{R}^{3 \times 4}$ is the modified error matrix from (4) consisting of the scale factor matrix $\mathbf{M}$ and the added bias vector $\boldsymbol{b}_g$ stacked together as follows:

$$\mathbf{Z} = \begin{bmatrix} s_x & 0 & 0 & b_x \\ 0 & s_y & 0 & b_y \\ 0 & 0 & s_z & b_z \end{bmatrix} \tag{6}$$

Note that matrix $\mathbf{Z}$ contains six unknowns. Additionally, (5) describes a single gyroscope position out of six. Each position is recorded for a period of $T$ seconds. By taking the average over the time we reduce the sensor noise. By repeating these two steps for each axis in each rotation direction we get:

$$\overline{\boldsymbol{\omega}}_{measured} := \begin{bmatrix} \overline{\omega}_{x^+} \\ \overline{\omega}_{x^-} \\ \overline{\omega}_{y^+} \\ \overline{\omega}_{y^-} \\ \overline{\omega}_{z^+} \\ \overline{\omega}_{z^-} \end{bmatrix}^T = \mathbf{Z} \cdot \begin{bmatrix} \omega_{x^+} & 1 \\ \omega_{x^-} & 1 \\ \omega_{y^+} & 1 \\ \omega_{y^-} & 1 \\ \omega_{z^+} & 1 \\ \omega_{z^-} & 1 \end{bmatrix}^T = \mathbf{Z} \cdot \boldsymbol{\omega}_{GT} \tag{7}$$

where $\overline{\boldsymbol{\omega}}_{measured}$ is a $\mathbb{R}^{3 \times 6}$ matrix of the averaged measured values, where each line represents a different axis at a different direction (one position), and the term $\boldsymbol{\omega}_{GT} \in \mathbb{R}^{4 \times 6}$ is the corresponding true angular velocity stacked. By applying the pseudo-inverse on (7) the unknown error terms comprising $\mathbf{Z}$ can be estimated:

$$\mathbf{Z} = \overline{\boldsymbol{\omega}}_{measured} \boldsymbol{\omega}_{GT}^{\mathrm{T}} \left( \boldsymbol{\omega}_{GT} \boldsymbol{\omega}_{GT}^{\mathrm{T}} \right)^{-1} \tag{8}$$

For a single axis (7) is simplified into two equations with two unknowns (bias and scale):

$$s_z = \frac{\overline{\omega}_{z^-} - \overline{\omega}_{z^+} - 2\omega_z}{2\omega_z} \tag{9}$$

$$b_z = \frac{\overline{\omega}_{z^+} + \overline{\omega}_{z^-}}{2} \tag{10}$$

where $\overline{\omega}_{z^+}$ is the averaged gyroscope z-axis while pointing up, $\overline{\omega}_{z^-}$ is the averaged gyroscope z-axis while pointing down, and $\omega_z$ is the GT angular velocity of the turntable. Solving (9)-(10) gives the gyroscope z-axis bias, $b_z$, and scale factor, $s_z$.

## 3. Data-Driven Calibration

This section describes our data-driven framework used to calibrate the gyroscope scale factor and bias. Figure 1 presents a block diagram of our proposed approach applied on the z-axis gyroscope. We design a simple, yet efficient, neural network for the bias and scale factor calibration as elaborated in the following subsections.

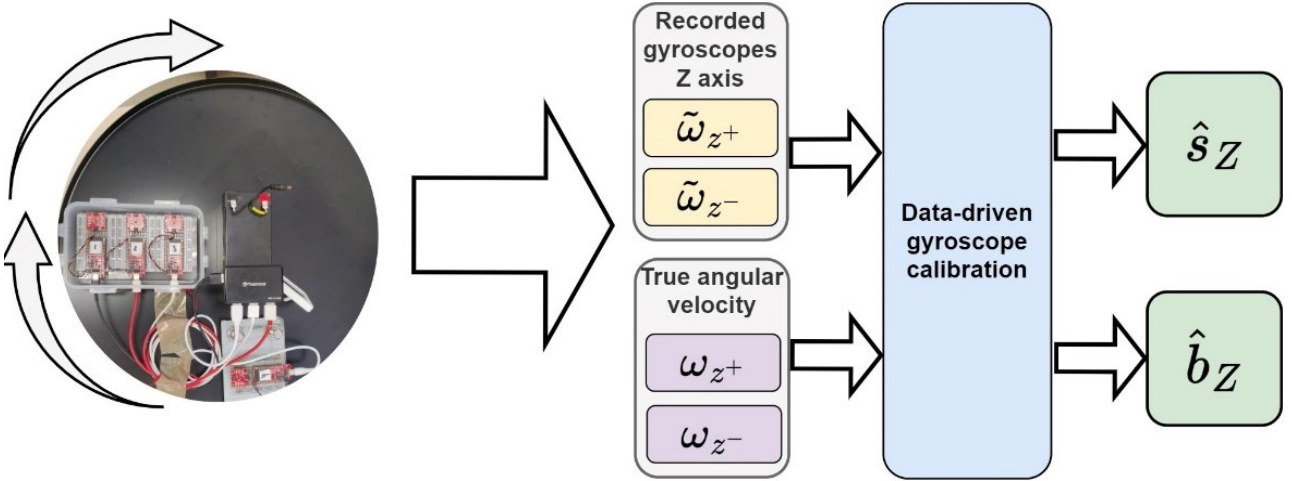

Figure 1. Block diagram illustrating our proposed approach on the gyroscope z-axis.

### 3.1. Neural Network Architecture

Our proposed data-driven framework is comprised of a 2-dimensional convolution neural network (2DCNN) head, followed by a fully connected (FC) block. The input to the network is the same as the input to the model-based approach, which is the recordings of the z-axis in both rotation direction, $\begin{bmatrix} \widetilde{\omega}_{z^+,t} \\ \widetilde{\omega}_{z^-,t} \end{bmatrix} \in \mathbb{R}^{2 \times T}$ with $t \in [1, \ldots, T]$, and the ground truth (GT) angular velocity, $\omega_{z,t} \in R_T$ as illustrated in Figure 1.

Our proposed approach architecture is a referred to as a multi-head architecture since the 2DCNN head is comprised of two separate 2DCNN heads which process separately each rotation orientation measurements against the GT values. The first 2DCNN head, referred to as the "Up head", process simultaneously both the clockwise and GT measurements as input, where the z-axis points upwards. The second head, referred to as "Down head", process the counterclockwise and GT measurements where the z-axis points downwards. The motivation for employing a multi-head architecture are 1) by using a 2D convolution kernel in each head, we promote the network to extract features and information from both the measurements and the GT once it enters the network, and 2) the use of a two separate 2DCNN heads is motivated by the model-based approach as presented in (7) where each measured rotation direction is connected by the error terms to the GT values.

Both 2DCNN heads are constructed similarly: first, batch normalization [21] layer, then 2D convolution kernel, followed by average pooling, and a leaky rectified linear unit (LeakyReLU) [22] activation function. A 2D convolution kernel is used to enable our proposed approach to simultaneously extract features from the inputs, similarly to [23] where we extracted information in a similar way from two input velocity vectors. Once the inputs are process by the two heads, the outputs are concatenated together to form a single input, which is then processed by an additional 2D convolution head. This combined 2DCNN head purpose is to further extract information and features, and its constructed as follows: a batch norm layer, followed by a 2D convolution layer, followed by average pooling, and then a LeayReLU activation is applied.

The output of the combined head, is flatted to form a vector, and fed to the FC block, which consists of two FC layers with a hyperbolic tangent (Tanh) [22] activation function followed by a Dropout [24] layer in between them. The output of the second FC layer is the estimated scale factor and bias of the z-axis. Figure 2 shows a detailed block diagram of the proposed data-driven architecture.

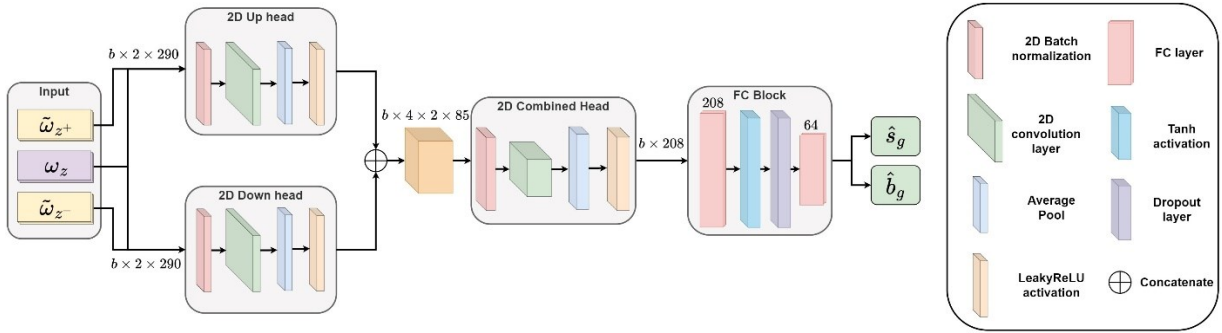

Figure 2. Block diagram showing the network's multi-head architecture, the input is divided and fed into the two 2D convolution heads, then into the combined 2D convolution head. Finally, the processed data is fed through a FC head to output the estimated gyroscope error terms.

In Figure 2 the input size to the two convolution heads is $b \times 2 \times w$, where $b$ is the batch size, 2 is the measured and GT angular velocity vectors stacked, and $w$ is the window size of the recordings. For illustration purposes we chose a window size of 290 samples.

## 3.2. Mathematical Formulation

Each fully connected layer is comprised of several neuron stacked together to process an input, commonly, each neuron in each layer is connected to each neuron in the previous and following layer, a single neuron in a fully connected layer is formulated as follows [25]:

$$z_i^l = \sum_{j=1}^{n_{l-1}} w_{i,j}^l \cdot a_j^{l-1} + b_i^l \tag{11}$$

where $a_j^{l-1}$ is the input to the $l^{th}$ layer from the previous $l - 1^{th}$ layer, $n$ represents the number of neurons in the previous $l - 1^{th}$ layer, $w_{i,j}^l$ is the learnable weight of the $i^{th}$ neuron in the current $l^{th}$ layer, that is connected to the $j^{th}$ neuron in the $l - 1^{th}$ layer, $b_i^l$ is the bias of a the $i^{th}$ neuron in the $l^{th}$ layer, and $z_i^l$ is the output of the neuron, which is used as input to the neurons it is connected to in the next $l + 1^{th}$ layer, and so on.

A convolution layer is comprised of a single or several kernels. A kernel is a $n \times m$ sized matrix where each input is a neuron. if $n = 1$ then the convolution layer is regarded as a 1D convolution, otherwise if $n \geq 2$ the convolution layer is referred to as a 2D convolution, as in our case. The convolution kernel performs a dot operation on the input as follows [25]:

$$c_{i,j}^l = \sum_{\alpha=0}^{n} \sum_{\beta=0}^{m} w_{\alpha,\beta}^r \cdot a_{(i+\alpha),(i+\beta)}^{l-1} + b_{\alpha,\beta}^r \tag{12}$$

where $c_{i,j}^l$ is the kernels output at the $l^{th}$ convolution layer at position $i, j$, the term $w_{\alpha,\beta}^r$ is the weight of the neuron at position $\alpha, \beta$ of the $r^{th}$ $kernel$, and $b_{\alpha,\beta}^r$ is the neuron bias, $a_{(i+\alpha),(i+\beta)}^{l-1}$ in the previous $l - 1^{th}$ layers output, which is the input to this convolution layer.

Commonly, a nonlinear activation function is applied in between each two connected neurons to prevent the neural network acting as a large linear regressor [22]. We apply two types of activation functions, the LeakyReLU and Tanh.

- LeakyReLU is applied over the output of each 2D convolution layer in all 2D heads, its formulation is as follows [22]:

$$LeakyReLU(c_{i,j}^l) = \begin{cases} c_{i,j}^l, & if\ c_{i,j}^l \geq 0 \\ \alpha \cdot c_{i,j}^l, & otherwise \end{cases} \tag{13}$$

where $\alpha$ is the scaling parameters which is set in advance and multiplies the convolution layers output $c_{i,j}^l$ as presented in (12).

- Tanh is applied over the output neurons in the first FC layer as follows [22]:

$$Tanh(z_i^l) = \frac{e^{z_i^l} - e^{-z_i^l}}{e^{z_i^l} + e^{-z_i^l}} \quad (14)$$

where $e \approx 2.71$ is the Euler number and $z_i^l$ is the FC layer's neuron output as presented in (11).

Batch normalization is commonly applied in deep learning to increase convergence speed and overall benefit the training process of networks and it does so by normalizing the input to intermediate layers in the network by learnable parameters [21]. In this paper we normalize the input to the "Up", "Down" and "combined" 2DCNN heads. The batch normalization formulation is as follows [21]:

$$C_{c,i,j}^l = \frac{I_{c,i,j}^l - \mu_c}{\sqrt{\sigma_c^2 + \epsilon}} \gamma_c + \beta_c \quad (15)$$

where $c$ is the number of channels in the input $I_{c,i,j}^l$, the terms $\mu_c$ and $\sigma_c^2$ are the mean value and variance of the whole input in channel $c$, the term $\epsilon$ is added for numerical stability, the terms $\gamma_c$ and $\beta_c$ are the parameters learned by the network during training for each input channel $c$. Note the terms $\mu_c$ and $\sigma_c^2$ are calculated for each input that arrives in the network, while $\gamma_c$ and $\beta_c$ are learnable parameters which are optimized each step in the training procedure.

Average pooling (AP) is applied to reduce the sparsity of the convolution layer output thus reducing training time, and extract meaningful features [26]. The AP formulation is as follows [27]:

$$\overline{C}_{i',j'}^l = \sum_{k=1}^{n} \sum_{d=1}^{m} \frac{1}{n \times m} (C_{k,d}^l), stride = (n, m) \quad (16)$$

where $n \times m$ is the size of the pooling kernel, $C_{k,d}^l$ is the input to the AP kernel and $\overline{C}_{i',j'}^l$ is the output. Notice that $i'$ and $j'$ are the new dimensions after AP operation. The AP layer is performed on all $c$ channels of the input $C_{k,d}^l$.

## 3.3. Multi-Task Learning and Training

Our proposed data-driven approach is designed to estimate two error terms simultaneously, the scale factor and bias of the gyroscope's z-axis. The purpose of the neural network training process is to minimize a "loss" function, by adjusting the weights and biases of the neurons (11)–(12). By deriving the loss function output with respect to said parameters, in a process known as back propagation, the networks parameters are adjusted and the loss value should decrease in the following training epoch. The loss function employed in this work is the root mean squared error (RMSE) scaled by a factor of 100 to promote generalization and defined as follows:

$$RMSE(\hat{y}, y_{GT}) = 100 \cdot \sqrt{\frac{1}{n} \sum_{i=1}^{n} (\hat{y}_i - y_{GT,i})^2} \qquad (17)$$

where $\hat{y}_i = [\hat{s}_i, \hat{b}_i]^T$ is the estimated error terms vector, and $y_{GT,i} = [s_{GT,i}, b_{GT,i}]$, is the GT error terms vector, both at index $i \in [1, ..., n]$. Therefore, this loss function is constructed such that it computes the joint loss for both estimated error terms and outputs a value that reflects the networks prediction error for both error terms.

Our proposed data-driven architecture is designed to estimate both the error terms in a single forward pass of the network, in other words, our approach estimates the scale factor and bias of the gyroscope's z-axis simultaneously. In deep learning literature, a neural network architecture designed to solve several tasks simultaneously is referred to as multi-task learning (MTL) [28]. MTL is a training paradigm that proposes that by designing a network in such a way that is solves two or more related tasks simultaneously, it might benefit the networks predictive capabilities. By learning the joint representations for the tasks, thus promoting faster learning and better results. MTL argues that learning related tasks together benefits the learning process more than training on each task separately.

From an inertial perspective, our proposed approach is designed to mimic the baseline approach with the same input and output and thus providing a competitive data-driven approach.

## 4. Experiment and Results

This section is comprised of the experiment description and collected data overview for this work is presented, followed by the results of the proposed approach evaluation on the recordings.

### 4.1. Experimental Setup

To evaluate our proposed approach, we designed an experiment campaign to record a constant angular velocity motion using a low-grade MEMS IMUs. The IMUs were attached to a turntable rotating at constant angular velocity of 78 degrees per second (DPS). The recordings were done once with the gyroscope's z-axis pointing upwards (clockwise rotation) and once with the gyroscope's z-axis pointing downwards (counterclockwise rotation). We attached the four IMUs on two different plates, three of them where attached to one plate and the fourth to a second plate, as shown in Figure 3. Both plates where mounted on the Vure-More VS BLDC turntable powered by Oriental Motor GFV2G200AS motor. The two plates holding the four IMUs were attached to a circular plate with a power source and micro-controller in the middle. Figure 3 shows an illustration of the experiment setup.

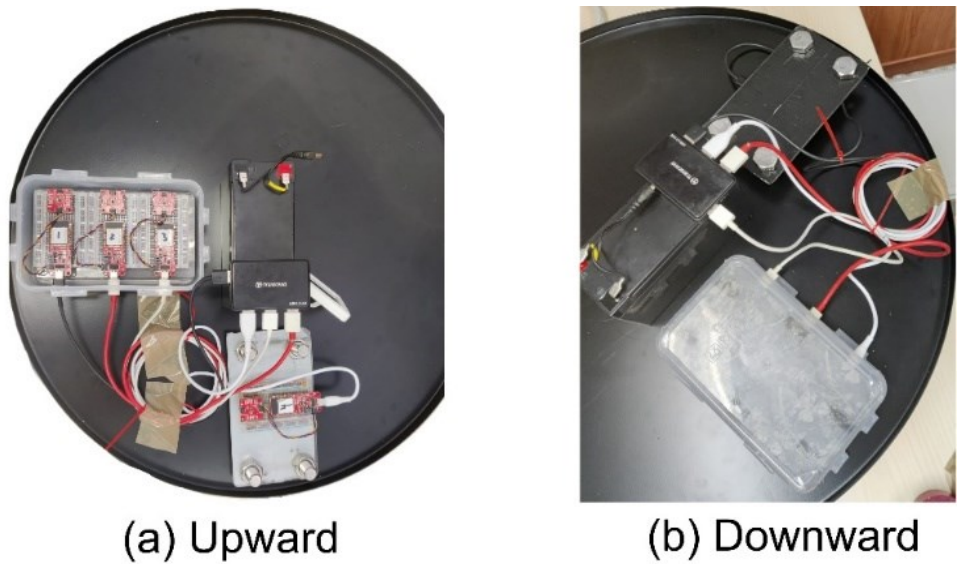

(a) Upward  (b) Downward

Figure 3: Images of our experiment setup showing the two plates with four IMUs, micro-controller, and the power source. (a) All IMUs are pointing upwards. (b) All IMUs are pointing downwards.

The MEMS gyroscope employed are the SparkFun LSM6DSO IMU, which were controlled by a SparkFun ESP32 micro-controller. The micro-controller sampled the IMUs at a sampling rate of 145 Hz. The IMU specifications according to the IMU datasheet are: bias is $\pm 2$ DPS, and the noise density is 3.8 DPS/$\sqrt{Hz}$.

*4.2.  Dataset*

 We performed three separate experiments in different days with different room temperate (uncontrolled). In the first experiment 13 scenarios were recorded. The second experiment consists of 7 scenarios while in the third experiment 28 scenarios was made. Each scenario was recorded in the following manner: first, a 70 seconds recordings was made while the z-axis pointed upwards. Then, the recording was paused, the plates were flipped such that the z-axis was pointed downwards, and an additional 70 seconds were recorded. After each scenario all the IMU's were shut down for 10 seconds and powered back on again to record the following scenarios. In total, we recorded 48 scenarios with a duration of 3.7 hours for all four IMUs.

Our recordings were 70 seconds long since in our preliminary testing this was found to be the required time for the baseline approach to converge. Additionally, we used here only the data from a single IMU, thus our dataset has 56 minutes of recordings from a single IMU.

We split our dataset into two: training set containing 46 recordings from all three experiments and a test set of two recordings from the same experiment, the last experiment, that were not included in the train set, referred to as TS1 and TS2. To generate the GT values of the desired scale factors and biases of each scenario, we performed the baseline approach calibration over the whole 70 seconds scenarios. These estimated scale factor and biases are regarded as the GT and were used during training and evaluation of our proposed approach. In the evaluation, calibration was made for a maximum time duration of six seconds to ensure that the model-based approach does not coverage.

To create a larger training set we split each scenario in the train set into six seconds long recordings. We split only the first 50 seconds as we observed that using a longer time period did not produce better results. Moreover, the test set scenarios scale factors and biases distribution might, and should, differ slightly from the training set values distribution, therefore we wanted to avoid overfitting and examine the generalization of our approach.

Next, each six second recording was split into two second windows, which are the input data points for the proposed approach (290 samples), with a 60% stride, resulted in nine windows. Each data point with a size of $2 \times 290$, where two is the input size.  Additionally, the GT angular velocity vector is stacked below to mimic the input to the baseline approach, thus each datapoint is now $3 \times 290$, where the first and second inputs are the two rotational directions measurements, and the third is duplicated vector of the true angular velocity, that is 78 DPS. In total, the training set is comprised of 1472 data points, which were split to train and validation by an 80:20% split ratio, resulting in 1177 data points for training and 295 for the validation. The labels for the train set are sized $1177 \times 2$, and $295 \times 2$ for validation. In each datapoint label, the first input is the scale factor, and the second is the bias. Figure 4 shows a histogram of the 1177 points of the train set.

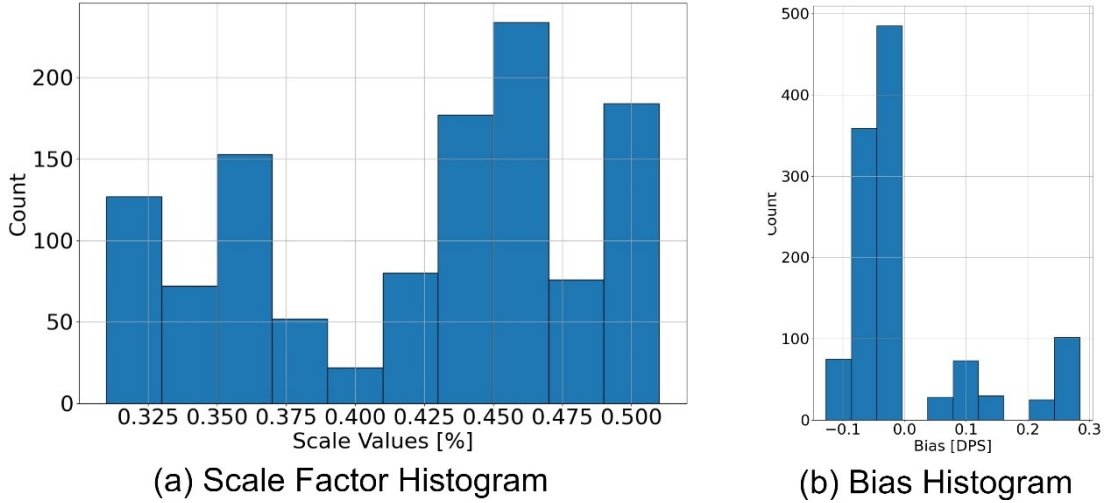

(a) Scale Factor Histogram  (b) Bias Histogram

Figure 4: Train set labels histogram. (a) Scale factor in % (b) Bias values in DPS.

### 4.3. Results

This section describes the calibration results of our proposed approach in comparison to the baseline approach. To evaluate our proposed approach, we look at the tradeoff between calibration accuracy and calibration time. To this end, we examined three calibration periods of 2, 4, and 6 seconds. Then, each approach estimated the error terms for each of the three calibration time periods for both TS1 and TS2. For each estimated error term in each calibration time, the following metrics were calculated:

1. The absolute error (AE):

$$AE_t = |\hat{y} - y_{GT}|, \qquad \forall t \in 2,4,6 \quad [seconds] \tag{18}$$

where $\hat{y}$ is the estimated error term (scale factor or bias), $y_{GT}$ is the GT error term, $t$ is the calibration window.

2. AE improvement of our approach over the baseline approach.

3. The calibration time improvement.

[13]

Table 1 shows the GT values of the scale factor and bias for both test scenarios. Those values where obtained by applying the model-based approach on the entire 70 seconds recordings.

Table 1: GT values of the scale factor and bias of the recordings in the test dataset.

|  | Scale Factor [%] | Bias [DPS] |
|---|---|---|
| **TS1** | 0.388 | -0.03073 |
| **TS2** | 0.414 | -0.07337 |

Table 2 shows the AE for the proposed approach and the baseline approach. From the table it is observed that our approach achieved accurate calibration results for all evaluated calibration times over both test scenarios. Moreover, even in 2 seconds calibration time, we observe an average improvement of 71% for the scale factor estimation over both TS1 and TS2, and an average of 51% improvement in the bias estimation of our approach over the baseline.

Table 2: The AE results for both approaches for each calibration window. Column "Improv [%]" shows the improvement of AE of our approach over the baseline for both error terms.

| Test scenario | Calibration window [sec] | Scale Factor | | | Bias [DPS] | | |
|---|---|---|---|---|---|---|---|
|  |  | Our approach | Baseline | Improv. [%] | Our approach | Baseline | Improv. [%] |
| **TS1** | 2 | 0.00015 | 0.00053 | 71.7 | 0.02195 | 0.03408 | 35.6 |
|  | 4 | 0.00013 | 0.00025 | 48.0 | 0.00283 | 0.00335 | 15.5 |
|  | 6 | 0.00008 | 0.00028 | 71.4 | 0.00703 | 0.01697 | 58.6 |
| **TS2** | 2 | 0.00024 | 0.00084 | 71.4 | 0.02615 | 0.07856 | 66.7 |
|  | 4 | 0.00023 | 0.00083 | 72.3 | 0.01272 | 0.05125 | 75.2 |
|  | 6 | 0.00013 | 0.00069 | 81.2 | 0.00572 | 0.02877 | 80.1 |

To test the calibration time improvement, we first calculated the baseline AE of both error terms at discrete intervals of one second increments for the total 70 seconds scenario, then we tested our proposed approach AE result at each calibration windows. Figure 5 shows the scale factor and bias AE values as function of time for both approaches.

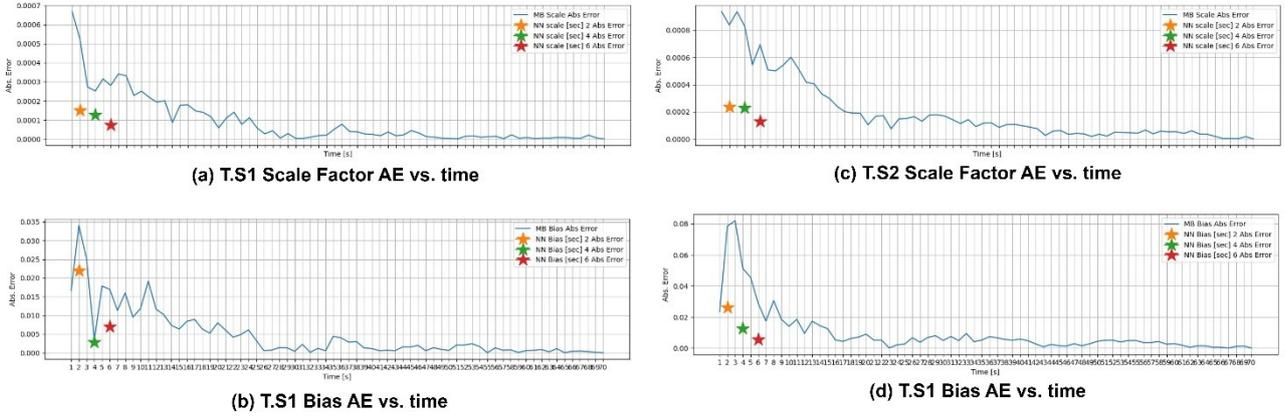

Figure 5: The AE results of our approach and the baseline for both error terms in both scenarios. The orange, green, and red stars represent the AE results for the 2, 4, and 6 seconds calibration windows of our proposed approach, respectively, and the blue line represents the baseline AE.

From Figure 5 it is evident that our proposed approach outperforms the baseline approach in the calibration windows. To estimate the convergence time improvement, we defined $T_{conv}$ as the time step $t \in [1,2,...,70]$ at which the baseline approach's AE, which is calculated for each $t$, achieves equal or lower AE value of our proposed approach. Notice that 1) $T_{conv}$ refers to the baseline approach, 2) it is determined for each AE based on the error terms estimated in each of the three calibration windows by our approach. For example, for TS1 our approach achieved an AE of 0.00015 after 2 seconds of calibration time, and the baseline approach achieved an AE of 0.00015 only after 17 seconds. Therefore, $T_{conv} = 17$, and our approach's convergence time improvement is 88%. Table 3 shows the $T_{conv}$ values achieved by the baseline approach for each of the three calibration windows evaluated by our approach, for both test scenarios.

From Table 3 we observe that for the two seconds calibration time, we achieve an average of 88% for the scale factor convergence time, and an average of 63% for the bias convergence time, over both TS1 and TS2. However, when considering the longest calibration time of six seconds, an average of improvement of 79% for the scale factor, and an average of 71% for the bias convergence times are achieved, averaging over both test scenarios. Overall, our approach achieves accuracy and convergence time improvements for all three test calibration windows over both test scenarios.

Table 3: Convergence time performance of our proposed approach based on the comparison of $T_{conv}$ against each calibration window evaluated by our approach for both TS1 and TS2.

| Test scenario | Calibration window [sec] | Scale Factor | | Bias [DPS] | |
|---|---|---|---|---|---|
| | | $T_{conv}$ | Improv. [%] | $T_{conv}$ | Improv. [%] |
| TS1 | 2 | 17 | 88.2 | 4 | 50 |
| | 4 | 23 | 82.6 | 26 | 84.6 |
| | 6 | 25 | 76.0 | 21 | 71.4 |
| TS2 | 2 | 17 | 88.2 | 9 | 77.8 |
| | 4 | 17 | 76.5 | 15 | 73.3 |
| | 6 | 34 | 82.4 | 21 | 71.4 |

## 5. Conclusions

Gyroscope measurements of MEMS-based IMUs are considered inaccurate due to dominant error terms and high sensitivity to the environment. Therefore, a gyroscope calibration procedure is carried out in order to mitigate those errors. Commonly, the six positions calibration approach is undertaken. This approach may require several minutes to converge. This work presents preliminary results for gyroscope z-axis calibration using a novel data-driven framework. Our proposed approach was trained and validated on a dataset of 56 minutes recorded using a low-performance MEMS-based gyroscopes. Our proposed approach achieved an average accuracy improvement of 72% for the error terms (bias and scale factor), while shortening calibration time by an average of 75%, compared to the baseline approach. Thus, shorter calibration times requiring only six seconds can be used. As a result of our research, it is possible to reduce the overall setup time before a mission while maintaining high accuracy. Furthermore, our approach can be adapted to estimate the error terms for all three axes, thus proposing an end-to-end calibration approach that is rapid.

## 6. Acknowledgment

This work was partially supported by data science research centre (DSRC), University of Haifa, Israel. Z.Y is supported by the Maurice Hatter Foundation and the presidential scholarship for outstanding direct track Ph.D. students.